# Smooth Path Planning with Subharmonic Artificial Potential Field

Bo Peng, Lingke Zhang, and Rong Xiong

*Abstract*— **When a mobile robot plans its path in an environment with obstacles using Artificial Potential Field (APF) strategy, it may fall into the local minimum point and fail to reach the goal. Also, the derivatives of APF will explode close to obstacles causing poor planning performance. To solve the problems, exponential functions are used to modify potential fields' formulas. The potential functions can be subharmonic when the distance between the robot and obstacles is above a predefined threshold. Subharmonic functions do not have local minimum and the derivatives of exponential functions increase mildly when the robot is close to obstacles, thus eliminate the problems in theory. Circular sampling technique is used to keep the robot outside a danger distance to obstacles and support the construction of subharmonic functions. Through simulations, it is proven that mobile robots can bypass local minimum points and construct a smooth path to reach the goal successfully by the proposed methods.**

## I. INTRODUCTION

Path planning for mobile robots is to find a safe and collision free path from the starting point to the goal in an environment with obstacles according to a certain evaluation criterion. It is widely used for obstacle avoidance in mobile robots, UAV, and underwater navigation scenarios. Artificial Potential Field (APF) methods have been widely used as both global and local path planners because of its simple principle, small amount of calculation, good real-time performance and path smoothness. However, traditional artificial potential field method has local minimum and the robot cannot reach the goal because of it. A lot of work have been done to solve this problem which can be divided into three main directions.

The first direction is to combine global information. Li et al. solve the local minimum and goal unreachability problems by changing the local repulsive potential field to a global repulsive potential field and using the concepts of safe distance and time distance detection methods [5]. Fan et al. combine Rapid Random Tree (RRT) algorithm with APF to generate global path first and use APF as a local planner [6]. But such methods can be time-consuming and computation-heavy for real-time planning, especially for embedded platforms.

The second direction is to add an additional vector or force at local minimum point to push the robot out of the area. Ping et al. add a small adjustment vector when the resultant force of the potential field on the robot is almost zero so the robot can jump out of the local minimum area [7]. Liu et al. add a k /2 angular offset to the robot's gravitational force when the combined force of the robot is below a threshold value [8]. Di et al. make the robot turn left according to the direction of the repulsive force to guide the robot out of the local minimum [9]. This kind of methods can keep the robot out of the local minimum point but cannot guarantee it will not fall in again. Thus, the path can be oscillating.

The third direction is to modify the potential functions' formulas to eliminate local minimum. Yuan and Shen add distance to the goal into the repulsive force. When the robot is detected to fall into a local minimum area, changing the power of the distance to the goal can have an impact on the resultant force and keep the robot move towards the goal [10]. Yuan adjusts the step length of planning and changes the forward angle when the robot is in the local minimum [11]. Zhang adopts an improved variable polynomial and increases the distance factor, which effectively solves the problems of unreachable targets and local minimum [15]. Jin-Oh et al. propose an elegant control strategy for real-time path planning by constructing harmonic functions using logarithmic functions. Harmonic functions can eliminate local minimum by their mathematical properties [1]. However, methods mentioned above did not consider the properties of the derivatives of modified potential functions' formulas. Usually, they can grow too rapidly when the distance to obstacle is too small so the path planning can be sawtooth-shaped.

In this paper, we use exponential functions to create subharmonic potential fields, which can help to eliminate the local minimum point and construct a smooth path for mobile robots in real-time. Meanwhile, a circular sampling technique is proposed to keep the robot outside a danger distance, which means that robots are free of collision outside this distance to obstacles, thereby avoiding collisions and further aiding the creation of subharmonic functions. In Section II, we introduce the properties of harmonic and subharmonic functions and provide proof of how our methods can effectively eliminate the local minimum point. In Section III, we discuss the blowing up of derivatives of other potential field functions when the robot is close to obstacles and show that our method will not have this problem. In Section IV, we introduce the circular sample technique and show how it can further support the elimination of local minimum and make sure the robot is outside a danger distance while heading to the target point. In Section V, we show the experimental results of how our proposed methods can construct a smooth path through obstacles and ablation testing results are presented to show the

Bo Peng is with the College of Control Science and Engineering, Zhejiang University, Hangzhou 310063, China. (corresponding author to provide phone: 18770026989; e-mail: percypeng5221@gmail.com).

Lingke Zhang is with the College of Control Science and Engineering, Zhejiang University, Hangzhou 310063, China (e-mail: zhanglingke49@gmail.com).

Rong Xiong is with the College of Control Science and Engineering, Zhejiang University, Hangzhou 310063, China. (e-mail: rxiong@zju.edu.cn)

validation of our method and how it can successfully eliminate the local minimum point. And we discuss the limitations of our approach and concludes in Section VI.

## II. Non-local Minimum of Subharmonic Functions

### A. Properties of Harmonic and Subharmonic Functions

A harmonic function is a twice continuously differentiable function f : U → R, where U is an open subset of $R^n$ that satisfies Laplace's equation, that is,

$$\frac{\partial^2 f}{\partial x_1^2} + \frac{\partial^2 f}{\partial x_2^2} + \cdots + \frac{\partial^2 f}{\partial x_n^2} = 0 \tag{1}$$

everywhere on U.

The first important property of a harmonic function is the principle of superposition, which follows from the linearity of the Laplace equation. That is, if $\varphi_1$ and $\varphi_2$ are harmonic, then:

$$a\varphi 1 + b\varphi 2, \quad a, b \in R \tag{2}$$

is also harmonic and a solution of the Laplace equation.

The second important property is the Maximum Principle: The maximum of a nonconstant harmonic function occurs on the boundary [1]. The same property can also be achieved for the minimum which is called the minimum property.

Subharmonic functions satisfy the following inequality:

$$\Delta f \geq 0 \tag{3}$$

and only the minimum property exists.

### B. Construction of Subharmonic Potential ields

Artificial potential field method plans the next move point from the distance to obstacles and goals. The magnitude of the forces is the same if the distance to obstacles or goals is the same. This means that the artificial potential field is spherical symmetrical. So it's natural to construct harmonic functions with spherical symmetry introduced from [2]. These functions depend only on $r$ (distance from the origin). To this aim, we can integrate the Laplace equation in a polar coordinate:

$$\nabla^2 \varphi = \frac{\partial^2 \varphi}{\partial r^2} + \frac{1}{r}\frac{\partial \varphi}{\partial r} = 0 \tag{4}$$

to get a solution [1]:

$$\varphi = c_1 \ln r + c_2 \tag{5}$$

The logarithmic function constructed to eliminate local minimum is used to improve the performance of artificial potential field method and achieved remarkable results [1]. The formulations would look like this:

$$\varphi_1(x, y) = \log_{10} r \tag{6}$$

where $r$ is the distance to the nearest obstacle:

$$r^2 = (x_r - x_o)^2 + (y_r - y_o)^2 \tag{7}$$

and $(x_r, y_r)$ is the position of the robot and $(x_o, y_o)$ is the position of the closet obstacle.

The relationship between traditional repelling potential and the distance to the closest obstacle is this [12]:

$$\varphi_2(x, y) = -\frac{1}{r} \tag{8}$$

Let's look at this example to see how local minimum is eliminated by harmonic functions. Suppose that there are four-point-obstacles at (1, 1), (-1, -1), (-1, 1), (1, -1). We use:

$$\varphi_1(x, y) = -\log_{10} r_1 - \log_{10} r_2 \\ -\log_{10} r_3 - \log_{10} r_4 \tag{9}$$

as the example harmonic function and

$$\varphi_2(x, y) = -\frac{1}{r_1} - \frac{1}{r_2} - \frac{1}{r_3} - \frac{1}{r_4} \tag{10}$$

as the example non-harmonic function where r is the distance to the point-obstacles.

According to the superposition property of harmonic functions, linear combinations of harmonic functions are still harmonic functions. So, using four point-obstacles will not affect whether the potential field is harmonic or not.

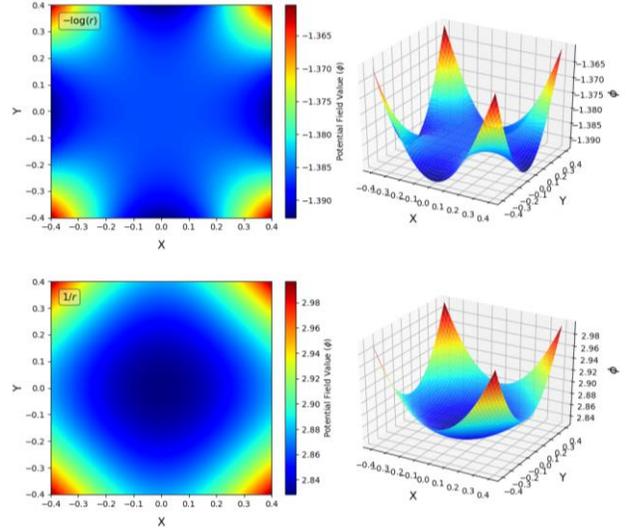

Figure 1. Potential field of harmonic and non-harmonic function.

Fig. 1 shows the artificial potential field of $\varphi_1$ (harmonic function) and $\varphi_2$ (non-harmonic function), respectively. There exists a local minimum at (0, 0) for the non-harmonic function but not for the harmonic function. The point (0, 0) is an unstable saddle point but not a local minimum.

Neither the harmonic functions nor non-harmonic functions aforementioned consider the properties of the derivatives of the functions, which have the form as:

$$\frac{\partial \varphi_1(x, y)}{\partial r} = \frac{\ln 10}{r} \tag{11}$$

$$\frac{\partial \varphi_2(x,y)}{\partial r} = \frac{1}{r^2} \qquad (12)$$

The derivatives near obstacles will grow much faster than points far from obstacles. Since the APF method calculates the next move point on the basis of the artificial forces generated by the potential field, rapid growth of the forces near obstacles will result in unstable and oscillating planning especially when there is a time lag between environment cognition and path planning. On the basis of the need for function formulations whose derivatives are more "moderate", we utilize exponential functions to meet the requirements.

The new formulation of the potential function would look like this:

$$\varphi_3(x,y) = -ae^{-kr^2} \qquad (13)$$

Now let's prove that $\varphi_3$ is a subharmonic function. Define function:

$$f(x,y) = r^2 = x^2 + y^2 \qquad (14)$$

$$\frac{\partial^2}{\partial x^2} e^{f(x,y)} = e^{f(x,y)} \left[ \left(\frac{\partial f}{\partial x}\right)^2 + \frac{\partial^2 f}{\partial x^2} \right] \qquad (15)$$

$$\frac{\partial^2}{\partial y^2} e^{f(x,y)} = e^{f(x,y)} \left[ \left(\frac{\partial f}{\partial y}\right)^2 + \frac{\partial^2 f}{\partial y^2} \right] \qquad (16)$$

and the full derivative is:

$$\Delta e^{f(x,y)} = e^{f(x,y)} \left[ \left(\frac{\partial f}{\partial x}\right)^2 + \left(\frac{\partial f}{\partial y}\right)^2 + \frac{\partial^2 f}{\partial y^2} + \frac{\partial^2 f}{\partial x^2} \right] \qquad (17)$$

Then we expand the formulation of function f(x) to get:

$$\Delta e^{x^2+y^2} = 4ke^{x^2+y^2}(kx^2 + ky^2 - 1) \ge 0 \qquad (18)$$

$\varphi_3$ is a subharmonic function and can eliminate local minimum outside the circle domain:

$$kx^2 + ky^2 \le 1 \qquad (19)$$

The radius of the circle domain is the danger distance ($d_0$)

$$\sqrt{\frac{1}{k}} = d_0 \qquad (20)$$

### C. Complete Potential Field Function

Exponential functions were used to model the repelling force of obstacles. We still need to construct an artificial potential field for the attract force of goals. The GNRON (Goal Not Reachable when Obstacles are Nearby) problem is identified by Ge and Cui in [3]. When an obstacle is close to the destination point, the robot's final resting position becomes shifted away from the destination point. This problem mainly happens because the attracting force is too small near the goal. Thus, we construct a subharmonic function whose derivatives are constant in the whole domain so the goal can stably attract the robot to it.

Let's assume that the goal is at the origin of the coordinate system and there is a point-obstacle at $(x_o, y_o)$.

We construct the attract potential field as:

$$\varphi_4(r) = a_{att} r^2, \quad a \ge 0 \qquad (21)$$

It also satisfies that:

$$\nabla^2 \varphi_4(r) = 4a_{att} \ge 0 \qquad (22)$$

where $a_{att}$ is a predefined parameter of the algorithm.

It is also a subharmonic function. Linear combinations of attract field and repel field ($-ae^{r^2}$) can still eliminate local minimum. The whole formulation of the potential field is:

$$\varphi(x_r, y_r) = \\ a_1(x_r^2 + y_r^2) - a_2 e^{-k_2[(x_r-x_o)^2 + (y_r-y_o)^2]} \qquad (23)$$

### III. SMOOTH PLANNING OF SUBHARMONIC FUNCTIONS

Use the same obstacles in Section II, we can plot the derivatives of the potential field functions.

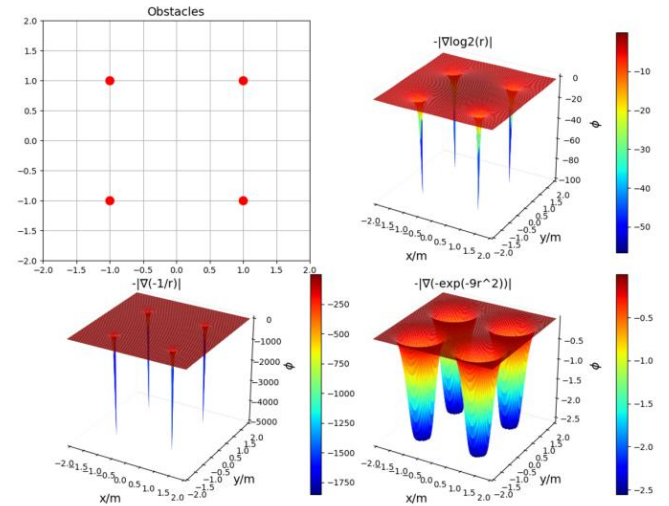

Figure 2. Derivatives of different potential field functions.

From Fig. 2 we can see that the values of functions (11) and (12) blow up when the robot is close to obstacles, but the value of function (13) does not. This means that when the robot is near the obstacles, the forces will become very big, and the next move point calculated from APF method will be very far.

To test these properties, we construct an environment where the distance to obstacles becomes smaller and smaller. In this simulation, we simply calculate the next move points based on the output of the artificial potential field method without considering the dynamics of robots. The results are shown in Fig. 3. We can see that the trajectories of traditional functions (8) and harmonic functions (6) keep oscillating until they plan a next move point across the obstacles which is unreachable to robots. However, the subharmonic functions can continue planning a smooth trajectory through thus supporting moderate properties of the derivatives of subharmonic functions.

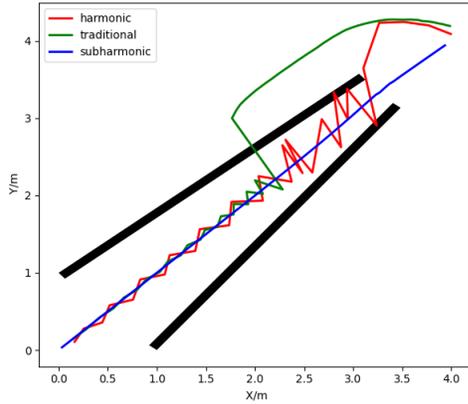

Figure 3. Trajectory of different potential field functions.

## IV. CIRCULAR SAMPLE TECHNIQUE

Subharmonic functions can eliminate local minimum, but the main disadvantage of using harmonic functions for path planning is when the robot is far away from obstacles, the derivatives of the functions has very small values, thus making a flat region where the robot can hardly move. Iniguez and Rosell used subharmonic functions for path planning in [13]. They used a 2d delta function to construct the attracting potential field by putting the goal position at the center of the delta function and model the obstacles as super-harmonic regions. And they solve the subharmonic function real-time based on the environment information using this formula:

$$u(r) = g(r, r_d) + \int_V \rho G dV_O + \oint_S (u \frac{\partial G}{\partial n} - G \frac{\partial u}{\partial n}) dS_0 \quad (24)$$

where:

- $u(r)$ is the potential field function, r is the position of the robot and $r_d$ is the position of the goal.
- $g(r, r_d)$ is the response to the delta function.
- $\int_V \rho G dV_O$ determines the influence of the sub- and super-harmonic regions, where $V_O$ is the volume of the obstacles.
- $\oint_S (u \frac{\partial G}{\partial n} - G \frac{\partial u}{\partial n}) dS_0$ represents the boundary conditions of the workspace.

In their method, the boundary conditions are not considered thus the third right hand term of equation 24 is 0. Outside the obstacles' region, the second right hand term is also 0 thus the potential field function is the only response to the delta function. However, the derivatives of the delta function can be very small when far away from the center.

To solve this problem, we propose a circular sample technique (Fig. 4). After we get the raw next move point from APF, we sample points on a circular line whose center is the next move point, and the radius is a predefined value (Fig. 4b). The points whose distance to obstacles is below the danger distance is left out for further process thus supporting inequality condition (19) for constructing subharmonic functions. Then we evaluate the generated points by using the APF function again to calculate the "next next move points". The closer the "next next move point" to the goal, the higher the generated point evaluation value (Algorithm 1). After that we add priority to a specific point, based on the position of the robot, obstacles and goal. In Fig. 5a, we find the point we want to add priority to clockwise along the circle along which we generated the points starting from the point closest to the goal. The first generated point whose distance to obstacles is above the danger distance will the point and its evaluation value will be increased a predefined value. When θ and d in Fig. 5 are both above certain values, we won't need to add priority to the generated points. If d is below a predefined threshold, we will start to add priority to certain points.

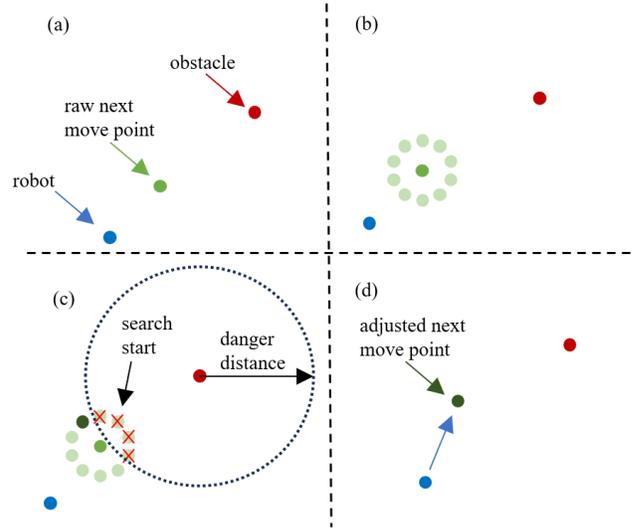

Figure 4. Circular sampling technique framework.

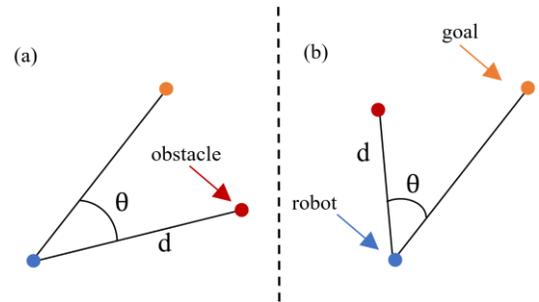

Figure 5. Add Priority Cases.

```
Algorithm 1: Evaluate points
  input : N Generated Points
  output: Evaluation result
1 generatedPoints is an array of N coordinates.
2 evaluationResult is an array of N numbers.
3 for i ← 0 to N − 1 do
4     for obs in pointObstacleArray do
5         if ||generatedPoints[i] − obs||_2 ≤ dangerDistance
6         then continueflag ← False
7             evaluationResult[i] ← ∞
8             break;
9         else continue;
10    if continueflag then
11        nextMovePoint ← APF(newSamplePoint);
12        evaluationResult[i] ← ||nextMovePoint − goal||_2
13    else continueflag ← True ;
```

## V. Ablation Testing Results

Here we perform our ablation testing using four different methods on four environments to prove the validity of the proposed algorithm and the elimination of local minimum point. We use the turtlebot simulation environment [14] with a burger robot on the ROS platform. The path planning frequency is 5 Hz.

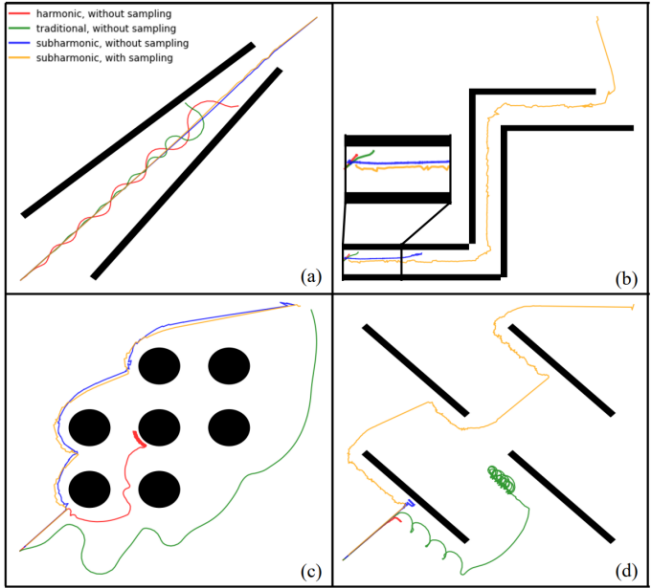

Figure 6. Ablation study result.

The trajectories of four different settings are shown in Fig. 6 (subharmonic functions $\varphi_3$ without sampling, subharmonic functions $\varphi_3$ with sampling, traditional functions $\varphi_2$ without sampling, harmonic functions $\varphi_1$ without sampling).

In Fig. 6a, we can see that subharmonic functions can plan a smooth path through narrow obstacles cause their derivatives are more "moderate". Other functions will keep oscillating until the robot hits obstacles. In Fig. 6b, we show that our proposed algorithm have the ability to successfully guide the robot through a tight traversable space, proving the robustness of our algorithm. In Fig. 6c and d, we construct different levels of local minimum. Fig. 6c has small local minimum located in front of the circle obstacle and Fig. 6d has large local minimum due to the big wall between the robot and the goal. Only our proposed method (subharmonic functions with sampling techniques) passed both small and large local minimum and successfully guide the robot to the goal, proving the elimination of local minimum of our method.

## VI. Conclusion

In this paper, we proposed a compact way to solve the local minimum problem by constructing subharmonic potential fields. Also, we proposed a circular sampling technique to make sure that the robot keeps a minimum danger distance to obstacles and the potential field is truly subharmonic. Experimental results have shown that there is significant improvement in solving local minimum problem. However, by using our algorithms, there are more parameters to handle than traditional potential field functions. Further work can be done to reduce the number of parameters tuning work by using other methods like fuzzy control based on the position and velocity of the robot and the distance to obstacles and goal.

## References

[1] J.-O. Kim and P. K. Khosla, "Real-time obstacle avoidance using harmonic potential functions," *IEEE Trans. Robot. Automat.*, vol. 8, no. 3, pp. 338–349, Jun. 1992.

[2] C. R. Chester, *Techniques in Partial Differential Equation*, New York: McGraw-Hill, 1971.

[3] S. S. Ge and Y. J. Cui, "New Potential Functions for Mobile Robot Path Planning," *IEEE Trans. Robot. Automat.*, vol. 16, no. 5, pp. 615-620, Oct. 2000.

[4] Q. Song and L. Liu, "Mobile robot path planning based on dynamic fuzzy artificial potential field method," *Int. J. Hybrid Inf. Technol.*, vol. 5, no. 4, pp. 85-94, Oct. 2012.

[5] Y. Li, B. Tian, Y. Yang, and C. Li, "Path planning of robot based on artificial potential field method," in Proc. 2022 IEEE 6th Information Technology and Mechatronics Engineering Conference (ITOEC), Mar. 2022, pp. 91-94.

[6] Q. Fan, G. Cui, Z. Zhao, and J. Shen, "Obstacle Avoidance for Microrobots in Simulated Vascular Environment Based on Combined Path Planning," *IEEE Robot. Autom. Lett.*, vol. 7, no. 4, pp. 9794-9801, Oct. 2022.

[7] S. Ping, L. Kejie, H. Xiaobing, and Q. Guangping, "Formation and obstacle-avoidance control for mobile swarm robots based on artificial potential field," in Proc. 2009 IEEE International Conference on Robotics and Biomimetics (ROBIO), Guilin, China, Dec. 2009, pp. 2273-2277.

[8] Q. Liu, J. Liu, Y. Zhao, R. Shen, L. Hou, and Y. Zhang, "Local path planning for multi-robot systems based on improved artificial potential field algorithm," in Proc. 2022 IEEE 5th Advanced Information Management, Communications, Electronic and Automation Control Conference (IMCEC), Chongqing, China, Dec. 2022, pp. 1540-1544.

[9] W. Di, L. Caihong, G. Na, S. Yong, G. Tengteng, and L. Guoming, "Local Path Planning of Mobile Robot Based on Artificial Potential Field," in Proc. 2020 39th Chinese Control Conference (CCC), Shenyang, China, Jul. 2020, pp. 3677-3682.

[10] J. Yuan and H. Shen, "Research on Local Path Planning of Mobile Robot Based on Artificial Potential Field Method," in Proc. 2019 IEEE 3rd Advanced Information Management, Communications, Electronic and Automation Control Conference (IMCEC), Chongqing, China, Oct. 2019, pp. 785-789.

[11] X. Yuan, "Research on the Limitations of UAV Path Planning Based on Artificial Potential Field Method," in Proc. 2022 9th International Forum on Electrical Engineering and Automation (IFEEA), Zhuhai, China, Nov. 2022, pp. 619-622.

[12] O. Khatib, "Real-time obstacle avoidance for manipulators and mobile robots," in Proc. 1985 IEEE International Conference on Robotics and Automation, St. Louis, MO, USA, Mar. 1985, pp. 500-505.

[13] P. Iniguez and J. Rosell, "Path planning using sub- and super-harmonic functions," in Proc. of the 40th International Symposium on Robotics, Mar. 2009, pp. 319-324.

[14] R. Amsters and P. Slaets, "Turtlebot 3 as a Robotics Education Platform," in *Lecture Notes in Networks and Systems*, vol. 111, 2020.

[15] W. Zhang, Y. Zeng, S. Wang, T. Wang, H. Li, K. Fei, X. Qiu, R. Jiang, and J. Li, "Research on the local path planning of an orchard mowing robot based on an elliptic repulsion scope boundary constraint potential field method," *Front. Plant Sci.*, vol. 14, pp. 1184352, Jul. 2023.